\documentclass[conference]{IEEEtran}
\IEEEoverridecommandlockouts
\usepackage{cite}
\usepackage{amsmath,amssymb,amsfonts}
\usepackage{algorithmic}
\usepackage{graphicx}
\usepackage{textcomp}
\usepackage{xcolor}
\def\BibTeX{{\rm B\kern-.05em{\sc i\kern-.025em b}\kern-.08em
    T\kern-.1667em\lower.7ex\hbox{E}\kern-.125emX}}
    
\usepackage[font=small,labelfont=bf]{caption}

\usepackage{color}

\newcommand{\eref}[1]{(\ref{#1})}
\newcommand{\sref}[1]{Sec. \ref{#1}}
\newcommand{\figref}[1]{Fig. \ref{#1}}
\DeclareMathOperator*{\argmax}{arg\,max}
\usepackage[normalem]{ulem}

\begin{document}

\title{On the Utility of Model Learning in HRI 
}

\author{
\IEEEauthorblockN{Gokul Swamy}
\IEEEauthorblockA{\textit{UC Berkeley} \\
gokul.swamy@berkeley.edu}
\and
\IEEEauthorblockN{Jens Schulz}
\IEEEauthorblockA{\textit{TUM, Germany} \\
jens.schulz@tum.de}
\and
\IEEEauthorblockN{Rohan Choudhury}
\IEEEauthorblockA{\textit{CalTech} \\
rchoudhury@caltech.edu}
\and
\IEEEauthorblockN{Dylan Hadfield-Menell}
\IEEEauthorblockA{\textit{UC Berkeley} \\ dhm@eecs.berkeley.edu}
\and
\IEEEauthorblockN{Anca D. Dragan}
 \IEEEauthorblockA{\textit{UC Berkeley} \\
 anca@berkeley.edu}
}
\maketitle

\begin{abstract}
Fundamental to robotics is the debate between model-based and model-free learning: should the robot build an explicit model of the world, or learn a policy directly? In the context of HRI, part of the world to be modeled is the human. One option is for the robot to treat the human as a black box and learn a policy for how they act directly. But it can also model the human as an agent, and rely on a ``theory of mind" to guide or bias the learning (grey box). We contribute a characterization of the performance of these methods for an autonomous driving task under the optimistic case of having an ideal theory of mind, as well as under different scenarios in which the assumptions behind the robot's theory of mind for the human are wrong, as they inevitably will be in practice. 
\end{abstract}


\section{Introduction}
An age-old debate that still animates the halls of computer science, robotics, neuroscience, and psychology departments is that between model-based and model-free (reinforcement) learning. Model-based methods work by building a model of the world -- dynamics  that tells an agent how the world state will change as a consequence of its actions -- and optimizing a cost or reward function under the learned model. In contrast, model-free methods never attempt to explicitly learn how the world works. Instead, the agent learns a policy directly from acting in the world and learning from what works and what does not. Model-free methods are appealing because the agent implicitly learns what it needs to know about the world, and \emph{only} what it needs. On the other hand, model-based methods are appealing because knowing how the world works might enable the agent to \emph{generalize} beyond its experience and possibly explain why a decision is the best one.

In neuro- and cognitive science, the debate is about which paradigm best describes human learning \cite{glascher2010states,lee2014neural}. On the other side of campus, in AI and robotics, the debate is instead about which paradigm enables an agent to perform its task best. As of today, model-free methods have produced many successes \cite{mnih2013playing,schulman2015trust,schulman2017proximal}, but some efforts are shifting towards model-based methods \cite{finn2016unsupervised,WinNT}.

\begin{figure}
    \centering
    \includegraphics[width=\columnwidth]{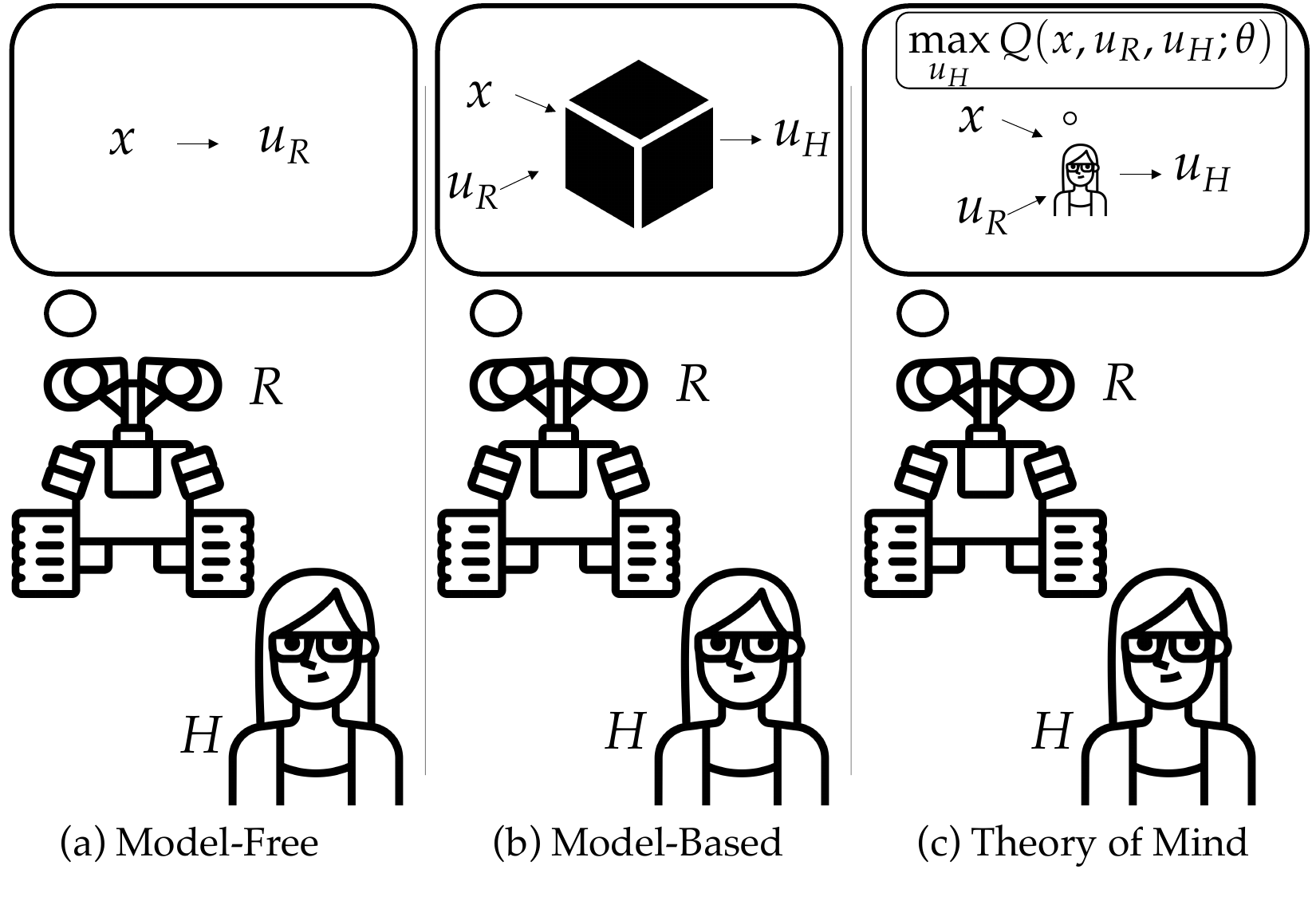}
    \caption{ We characterize the performance of three HRI paradigms that impose increasingly more structure: (a) in model-free, the robot learns a policy for how to act directly, without modeling the human; (b) in black-box model-based, the robot learns a policy for how the human acts, and uses it when optimizing its reward; (c) in theory of mind, the robot further assumes that the human optimizes a reward function with unknown parameters. }
    \label{fig:front}\vspace{-1em}
\end{figure}

In the context of Human-Robot Interaction (HRI), which is our focus in this work, the debate has a different nuance. For robots that do not work in isolation, but in worlds that contain people, the dynamics of the world are no longer just about how physical state changes, but also how \emph{human} state changes -- what the human will do next, and how that is influenced by the robot's action. There is thus a lot of richness in what it means to be model-based. On the one hand, the robot can learn a model by observing human state transitions and fitting a policy to them, as it can with any other part of the environment (\figref{fig:front},b). This is a ``black-box" approach to system identification. But on the other hand, the robot can structure its model and explicitly reason about the human differently: humans, unlike objects in the world, have \emph{agency}, and treating them as such means using what Gopnik called a ``Theory of Mind" (ToM) \cite{gopnik199410}: a set of assumptions about how another agent works, including how they decide on their actions (\figref{fig:front},c). Rather than black-box, this is a ``gray-box" approach. 

When it comes to our own interaction with other people, cognitive science research has amassed evidence that we might use such a theory of mind \cite{carruthers1996theories,gergely2003teleological,sodian2004infants,baker2009action} -- in particular, that we assume that others are approximately \emph{rational}, i.e. they tend to make the decisions that are approximately optimal under some objective (or utility, or reward) \cite{becker2013economic}. But even if humans do use such tools in interaction, it is not at all clear that robots ought to as well. For robots and interaction, methods from all three paradigms exist: model-free \cite{nikolaidis2013human,busch2017learning,reddy2018shared}, regular model-based \cite{schmerling2017multimodal}, and ToM-based \cite{ziebart2009planning,dragan2013legibility,bai2015intention,javdani2015shared,sadigh-human-rss2016}.

Ideally, we'd want to settle the debate for HRI by seeing what works best in practice. Unfortunately, several factors make it very difficult to get a definitive answer: 1) we do not yet have the best ToM assumptions we could get, as research in the psychological sciences and even economics is ongoing; as a result, any answer now is tied to the current models we have, not the ones we could have; 2) making the comparison requires immensely expensive evaluations with robots acting in the real world and failing in their interactions around real people -- this is very difficult especially for safety-critical tasks; 3) the answer might depend on the amount of data that can be available to the robot, and the distribution that data is sampled from, which is also hard to predict -- therefore, what we need to know is which paradigm to use as a function of the data available.

In this work, rather than attempting to answer the practical and more general question of which paradigm the robot should use, our idea was to turn to a more scientific and focused question. \emph{We do not know how wrong the eventual ToM model will still be, and we do not know how much data we can afford, but we can compare the performance of these paradigms under different possible scenarios.}

We take inspiration from recent work in learning for control \cite{tu2017least} that studied the sample complexity of model-based and model-free methods for a very simple system: the linear quadratic regulator. Their idea was that performance in a controlled simple system that we have ground truth for is informative -- if a method struggles even on this system, what chance does it have in the real world? 

We thus set up a simplified HRI system: we have a robot that needs to optimize a reward function in the presence of a human who optimizes theirs, in response to the robot's actions. We instantiate this in an autonomous driving domain. This simplification from the real world enables us to start answering two fundamental questions: 1) how big the benefit of ToM would actually be, even in the optimistic case of having the perfect set of assumptions about human decision making; and 2) how exactly this benefit decreases as our assumptions become increasingly incorrect. 

\noindent\textbf{Findings.} We uncover evidence which suggests that if we had a perfect theory of mind, we'd reduce the sample complexity of learning compared to black-box model-based learning, and, more importantly, we might be able to learn solely from off-policy data. In contrast, we found that it is possible for black-box model based approaches to fail to improve even with a lot of data, unless that data comes from the right distribution -- the off-policy method failed for us, and even iteratively collecting on-policy data did not seem to be enough in our system without targeted exploration. 
ToM was surprisingly robust to wrong assumptions: in our driving domain, even when the human deviates from ToM, it could produce useful predictions -- but this was in part due to ToM's wrong predictions interacting well with short-horizon planning, rather than ToM making better predictions than black-box approaches. 
Further, ToM tended to transfer better: when we used a trained model to interact in a new HRI system, the ToM model performed better even when the human deviated substantially from training behavior.

\noindent\textbf{Limitations.} These findings should be of course taken with a grain of salt. The differences we introduce between the ToM and the ground truth "human" might not be representative of what differences we will see in real life. We also do this for one particular task, with one particular ToM instantiation. Because of the difference between accurate predictions and \emph{useful} predictions, different tasks might lead to different results. In a previous iteration of this work, we investigated a task where a robot was attempting to influence the human driver from behind to move out of the way: we realized that most wrong predictions would lead to the correct plan in such a scenario, because most wrong predictions would lead to the human moving out of the robot's lane. We specifically designed the current task so that being able to capture the influence on the human would actually matter more to task performance. This is not true of all tasks.  
Further, the realities of planning in continuous state-action spaces even in our relatively simple scenario, such as limited planning horizons and convergence to local optima, put their own stamp on the findings.  That said, the results give us a glimpse at what might happen, pointing to surprisingly nuanced trade-offs between ToM and black-box approaches, in particular with respect to the type of data required and the ease of initialization.

Overall, we are excited to contribute a quantitative comparison of these paradigms for an HRI task, along with an analysis of their degradation as we make the wrong modeling assumptions, decrease the amount of data available, or restrict the ability to collect data on-policy and to explore.

\section{Interaction Learning Algorithms}
\subsection{Notation}
For all of the following methods and experiments, we denote the human plan at time $t$ by $\textbf{u}^t_H$, and the robot plan at time $t$ by $\textbf{u}_R^t$. We also denote the action executed by the human at time $t$ by $u^t_H$, and for the robot $u^t_R$. Both the human and robot states at time $t$ are denoted by $x_H$ and $x_R$. 

\subsection{Robot Objective}
The robot's goal is to optimize a reward function $r_R(x_R,x_H,u_R,u_H)$ that depends on both its state and action, as well as the human's state and action.

\subsection{Theory-of-Mind-Based Learning}\label{sec:tom}
In line with previous work  \cite{gergely2003teleological,sodian2004infants,baker2009action}, our ToM will assume that the human optimizes a reward function. The robot will focus the learning on figuring out this reward via inverse reinforcement learning, and the ToM-based method will plan the robot's actions using the learned model.

In particular, the reward will have the form
$$r_H(x_H, x_R, u_H, u_R;\theta) = \theta^T\phi(x_H, x_R, u_H, u_R)$$
where $\theta$ is a vector of weights and $\phi$ is a feature map from the current state of the system, which depends on the robot's state and action as well, akin to the robot's reward. We describe the particular features we assumed in more detail in \sref{sec:groundtruths0}. 

To optimize their reward, the person needs to reason about the inter-dependency between their future actions and those of the robot. Work on ToM has investigated different ways to capture this, from infinite regress (i.e. "I think about you thinking about me thinking about you...") to capping the regress to one or two levels.
Our particular instance of ToM is based on prior work which avoids regress by giving the human access to the robot's future plan \cite{sadigh-human-rss2016}, and is given by 
\begin{equation}
\textbf{u}_H^*(\textbf{u}_R;\theta)=\argmax_{\textbf{u}_H} \mathcal{R}_H(x_H,x_R,\textbf{u}_H,\textbf{u}_R;\theta)\label{eq:tom_opt}
\end{equation}
with $\mathcal{R}$ denoting the cumulative reward, i.e. the sum of rewards over a finite horizon of the length of the trajectories $\textbf{u}_H$ and $\textbf{u}_R$. This assumption is very strong, i.e. that the human can read the robot's mind. Part of our goal is to simulate the human as instantiating other approaches, and teasing out how useful or not ToM still is when its assumptions are wrong. 
This particular ToM instance will also assume that the person computes this at every step, takes the first action, observes the robot's new plan, and replans.

To leverage this ToM model, the robot needs to know the human reward function $r_H$.
In our experiments, we create a dataset of demonstrations $\mathcal{D}$ by placing our ground-truth human (be it a human perfectly matching the ToM model or one that does not) around another vehicle executing various trajectories, and recording the human's response. In the real world, this data would be collected from people driving in response to other people. 
We then run inverse reinforcement learning \cite{abbeel04inverse, levineCIOC, ziebart08maxent,  sadigh-human-rss2016} on $\mathcal{D}$ to obtain weights $\theta$.

Finally, given the human reward function described by the learned weights $\theta$, the robot optimizes its own plan:
$$\textbf{u}^*_R(\theta) = \argmax_{\textbf{u}_R}
\mathcal{R}_R(x_R,x_H,\textbf{u}_R,\textbf{u}_H^*(\textbf{u}_R;\theta))$$
with $\mathcal{R}_R$ denoting the robot's cumulative reward. The robot plans, takes the first action, observes the next human action, and replans at every step. We use a Quasi-Newton optimization method \cite{Andrew07L1Reg}
and implicit differentiation to solve this optimization problem.

The main challenge with the ToM approach is that in reality people will not act according to $\textbf{u}_H^*(\theta)$ for \emph{any} $\theta$. In general, ToM models will be misspecified, failing to perfectly capture human behavior.

\subsection{Black-Box Model-Based Learning}\label{sec:model-based}
The Theory-of-Mind approach models the human's actions as explicitly optimizing some cost function.  An alternative is to learn this function directly from data via, e.g., a conditional neural network, as in \cite{schmerling2017multimodal}. 


\noindent\textbf{Off-Policy (MB-OFF).}
In the ``off-policy" black-box model-based approach, we collect a training dataset $\mathcal{D}$ of human-robot interactions as in the ToM approach and 
fit a neural network $f$ to $\mathcal{D}$. This allows us to estimate the human plan $\textbf{u}_H$, given the human and robot histories, the current state of the system, and the robot plan: 
$$\textbf{u}_H = f(H_R, H_H, x_R, x_H, \textbf{u}_R).$$
$H_R$ and $H_H$ are the state-action histories of the robot and human over a finite time horizon. 
The model $f$ is trained by minimizing the loss between $f(.)$ and the observed data $\textbf{u}_H$ in $\mathcal{D}$. Specifically, we use a neural network with three fully connected layers, each with 128 neurons and ReLU activations, and train the network using ADAM \cite{adam}. 


Given this learned model, the vanilla model-based method generates a plan with trajectory optimization, just like the ToM method:
$$\textbf{u}^*_R(f) = \argmax_{\textbf{u}_R}
\mathcal{R}_R(x_R,x_H,\textbf{u}_R,f(H_R, H_H, x_R, x_H, \textbf{u}_R)).$$
The distinction is whether the prediction about $\textbf{u}_H$ comes from optimizing $\mathcal{R}_H$, or from the black box predictor $f$.

\noindent\textbf{On-Policy (MB-ON).}
The off-policy approach ignores the fact that the predictive model influences the behavior of the robot and therefore the future states on which the robot will query the predictor. In fact, if $f$ attains a test error rate of $\epsilon$ it can still exhibit prediction errors that are $O(\epsilon T^2)$ when rolled out over a horizon $T$~\cite{ross2011reduction}. The trajectories generated by optimizing against a fixed model induce \emph{covariate shift} that reduces the accuracy of the model. 


To deal with this, a typical approach in system identification is to alternate between fitting a model, and using it to act and collect more interaction data. This comes at a cost: the need for data collected on-policy, from interacting with the human, rather than from (off-policy/offline) demonstrations.


In this method, we again use a neural network $f$ to predict the human actions. But unlike the previous model-based method, which relies on training data collected beforehand, here we train $f$ \textit{iteratively}. 



Of course, needing interaction with the human can be prohibitive, especially if a) a lot of interaction data is needed, or b) the robot does not perform well initially, when its model is not yet good, which can harm adoption or lead to safety concerns.

\noindent\textbf{Idealized (MB-I).} 
Even collecting data on-policy has no guarantee that the robot will acquire a human model that is conducive to the \emph{optimal} robot policy, i.e. if we assume the human acts according to a function $f_{GT}$, the robot learning on-policy might not converge to $\textbf{u}^*_R(f_{GT})$. This is because the robot might not explore enough. However, getting exploration right is an open area of research. We thus also test a method that uses idealized exploration -- we collect data not only on-policy, but rather also "on-the-optimal-policy". This method in a sense "cheats" by getting access to data collected by acting according to $\textbf{u}^*_R(f_{GT})$ from different initial states. We then mix this data with on-policy data, iteratively, to obtain a model that can cover both the current and previous distributions, as well as the idealized distribution (we found that this mixing was important to the performance compared to using the idealized distribution alone). The reason this method is worth considering, despite the very strong assumption of access to the idealized distribution for "exploration", is that robots might be able to use human-human interaction data in situations where that optimal robot policy might coincide with what people do with each other.

\subsection{Model-Free Learning}\label{sec:model-free}
A final approach is to employ fully model free methods, such as policy gradients or DQNs. These methods are quite general and fast at inference time as they make no assumptions about the environment and don't explicitly plan online. We use Proximal Policy Optimization (PPO) \cite{schulman2017proximal}, a model free reinforcement learning algorithm that has had strong results in other continuous control tasks. Although more sample efficient approaches exist, we selected PPO because it has been adopted as a sort of baseline for continuous control tasks. 

PPO works by computing clipped gradients of expected reward with respect to the parameters of a policy. This gradient is estimated with rollouts using the current policy parameters in the environment. The algorithm alternates between rolling out trajectories and performing gradient updates. See \cite{schulman2017proximal} for a more complete explanation of the approach.  We used the PPO2 implementation in the Stable Baselines repository  \cite{baselines} with default parameters as our model-free algorithm.
The primary difficulty in applying PPO to this setting is that the human simulator we implemented (for testing what happens where ToM assumptions are exactly right), does not, strictly speaking, fit into the environment model used for reinforcement learning. Because the human reacts to what the robot \emph{plans} to do in the future, the environment is different depending on the current policy parameters. We implement this by having the human respond to a robot plan that was generated assuming the person continues at a constant velocity for the rest of the planning horizon. Because of how precise the controls need to be for this task, we do not sample from the distribution over actions the robot's policy outputs during evaluation time and instead use the mean. We also clip some of the penalties for catastrophic events like driving off the road or colliding with another vehicle and terminate the episode early to not poison the reward buffer of the policy gradient algorithm.






\section{Performance under \\Correct Modeling Assumptions}
We begin with comparing these methods in a driving domain.

\subsection{Experiment Design}
\label{sec:groundtruths0}

\noindent\textbf{(Ground Truth) Human Simulator.}
For our driving domain, states are tuples of the form $(x, y, v, \alpha)$, where $x$ and $y$ are the positional coordinates, $v$ the speed, and $\alpha$ the heading. The actions are $u = (a, \omega)$, where $a$ is a linear acceleration, and $\omega$ an angular velocity.

The ground truth human simulator plans forward over a finite time horizon $T$ (in all experiments, $T=5$) by optimizing over a linear combination of features:

\noindent\emph{Car Proximity}: This cost is based on the distance between the robot and human cars, which represents human's desire to not hit another vehicle. Given the human state $(x_H, y_H, v_H, \alpha_H)$ and the robot state $(x_R, y_R, v_R, \alpha_R)$, this cost is given by $\mathcal{N}((x_R, y_R)|(x_H, y_H, \sigma_{car}^2)$.

\noindent\emph{Lane Edge Proximity}: This cost is based on the distance to the nearest lane edge. This represents how humans generally prefer to stay in the middle of their lane. Letting the left edge of some lane be $L_l$ and the right edge $L_r$, the lane cost is given by: $\mathcal{N}(L_1 | x_t, \sigma_{lane}^2) + \mathcal{N}(L_r| x_t, \sigma_{lane}^2)$.

\noindent\emph{Forward Progress}: This cost is based on the vertical distance between the next state and the current state, representing how humans want to go forward when driving. This cost is given by: $-(y_{t+1} - y_t)$.

\noindent\emph{Bounded Control}: This cost is based on accelerating or trying to turn more quickly than certain bounds, representing how humans prefer smoother rides and cars have actuator limits. The cost is given by $\exp(a - a_{max}) + \exp(\omega - \omega_{max})$.

\noindent\emph{Offroad}: This cost represents how drivers want to stay on the road when driving. Letting the left edge of the road being $R_l$ and the right edge $R_r$, the offroad cost is given by $\exp(x_t - R_l) + \exp(R_r - x_t)$.



The weights on these features were tuned to produce plausible/natural driving in a series of scenarios.
In addition to the features and weights, the ground truth human simulator is given the plan of the robot $\mathbf{u}_R$. It then solves the cost minimization in \eref{eq:tom_opt}. Importantly, this particular simulator exactly matches the assumptions made by the ToM learner (Section \ref{sec:tom}), in both the features used and planning method. The next section modifies the simulator so that the ToM assumptions are wrong.

\noindent\textbf{Environment.}
The experiment environment consists of the human and robot car on a three-lane road, with the robot beginning in front of the person. The robot car starts in the middle lane while the human car starts in either the middle or left lane. The robot has a similar set of features to that of the human, incentivizing it to make progress, avoid collisions, keep off of the lane boundaries, and stay on the road. However, we add another feature representing the penalty associated with colliding into a pair of trucks that are in front of the robot and occupy the center and right lanes. 

We chose this as our environment because for all its simplicity, it can actually capture sophisticated interaction: given our ground truth human, to do well, the robot should merge in front of the person in the left lane by \emph{influencing the person to make space} rather than passively waiting for the person to pass by and accumulating significantly less reward for forward progress. We use this scenario as an interesting challenge for HRI, because it requires being able to account for robot's \emph{influence} on the human. Accordingly, we initialize the ToM and black-box model-based methods by pre-training them on a dataset in which the human drives in isolation -- this makes it so that both methods get access from the start to the basics of human driving (staying on the road, in the lane, accelerating to reach the speed limit), without giving away any of the aspects that are important to this particular task, i.e. the influence.


\begin{figure*}
    \centering
    \includegraphics[width=\textwidth]{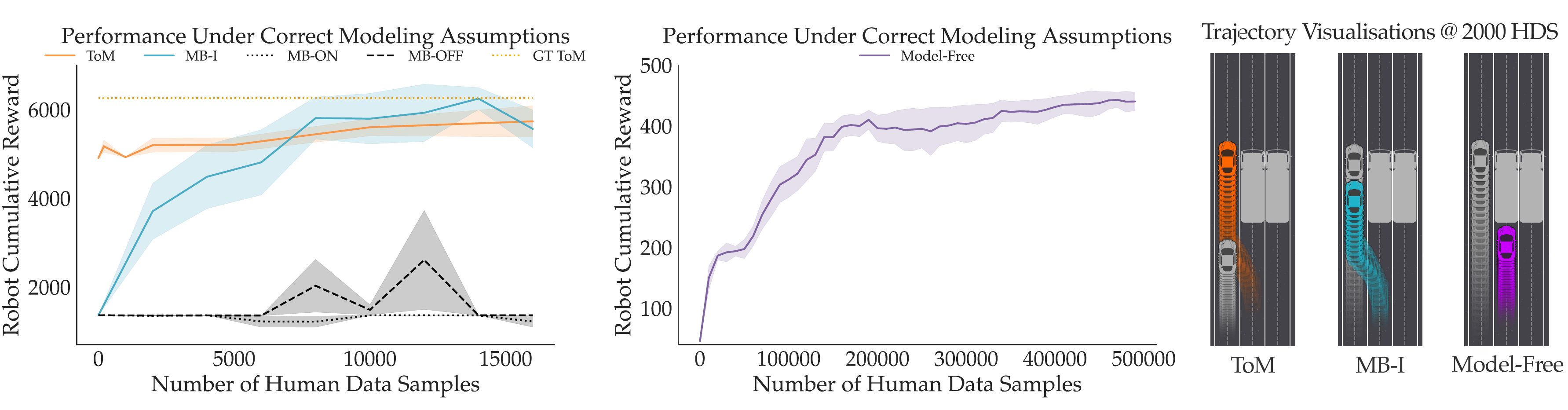}
    \caption{The test rewards of the interaction learning algorithms on the scenario with the ground truth human simulator. The ToM learner has the smallest sample complexity, followed by the MB-I method. The MB-ON and MB-OFF methods do not appear to improve from their original performance. The ToM is able to consistently merge in front of the human car most quickly, and thus obtains the highest reward, while on average, the MB-I method waits for the person to pass before merging.}
    \label{fig:s0}\vspace{-1em}
\end{figure*}
\begin{figure}
    \centering
    \includegraphics[width=.8\columnwidth]{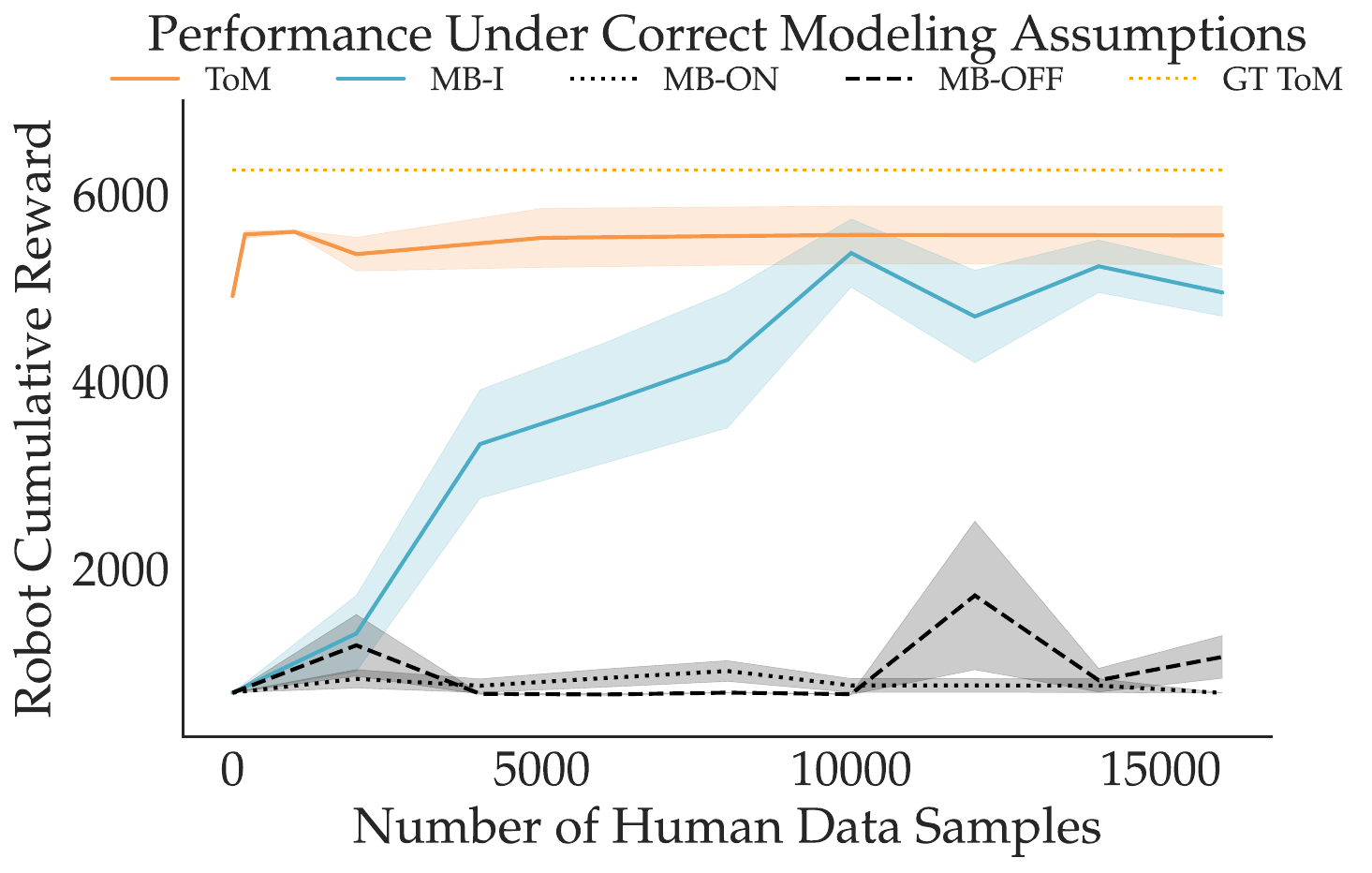}
    \caption{The comparison between ToM and black-box model-based for modification of our environment setup where there are no control bounds for the cars. This matches the result from \figref{fig:s0}}
    \label{fig:s0unbounded}
\end{figure}

\noindent\textbf{Manipulated Variables.}
We manipulate two variables: the interaction learning algorithm (with the options described in the previous section) and the amount of data (number of samples) the learner gets access to.\footnote{Note that the $u_R$ and $u_H$ collected in training data and used in learning algorithms are not \textit{plans}, but rather \textit{actions} that have been executed by the robot and human, since in reality, humans can only react to and learn from actions they can physically observe.} For the off-policy methods, we use data from a different type of interaction where the robot drives at constant velocity near the human. 

\noindent\textbf{Dependent Measures.}
In each experiment, we measure the reward the robot accumulates after training over test environments drawn from a more challenging subset of the training environment initial state distribution. 
We train ToM and model-based 5 times with each data sample size and measure test reward for each trained model. We train model-free only four times due to the several orders of magnitude larger amount of data required.

\subsection{Analysis}
We plot the results and visualize trajectories from each paradigm in \figref{fig:s0}.
We see that ToM starts with higher performance even at 0 samples. This is because while both models are pre-trained with a data from a human acting in isolation and thus make very similar predictions, the \emph{gradients} of those predictions with respect to the robot's plan are different, and that in turn leads to different plans. With the black-box model, the gradients are close to 0 due to the non-reactive training data. This results in the robot planning to let the human pass and merge behind. The ToM model, on the other hand, anticipates influence on the human's plan because of its structure (i.e. the collision avoidance feature), despite pre-training on non-reactive data (which settles on a nonzero weight for said feature). This enables the robot to figure out that it can merge in front of the human even from the start. This speaks to a certain extent to the ease of initializing ToM: because it encodes that people want to avoid collisions, and that human actions are influenced by robot actions, the original solution, albeit imperfect, is already qualitatively achieving the desired outcome.

As the ToM model gets off-policy data from the ground truth human simulator, the robot's reward when planning with ToM improves. Looking at a representative trajectory (right of \figref{fig:s0}), we see that the robot gets in front of the human, who slows down to accommodate. 

On the other hand, the off-policy model-based method, which learns from the same demonstrations used by the ToM approach, is unable to learn to merge in front of the human car. One explanation for this behavior is the difference between the training and test distributions -- at test time the robot plans with the learned model, generating different kinds of trajectories with different human responses than in training. Concretely, the off-policy data has a mild influence of the robot's actions on the human, whereas solving our test merge task requires being able to model stronger influence. The ToM model can better cope with this because it has access to the strong predefined structure various features in the reward function, and can rely on trajectory optimization to produce the corresponding human trajectories in new situations, given new robot trajectories. 

What is more surprising at a first glance, however, is that the on-policy method also fails to improve. This points to a potentially fundamental problem with model-based methods in this type of interactive situation: they start not knowing how to influence the human, so the initial policy does not attempt to merge in front and instead lets the human pass. Then, they collect data on-policy, but none of the roll-outs end up influencing the human's behavior enough, so their model never evolves to capture that potential stronger influence. As a result, the robot's policy never attempts the merge and remains at a low reward.

The idealized model-based, which "explores" by getting access to data from the optimal policy in response to the \emph{ground truth} human model, does get up to the ToM performance, but requires more data. We also see these results reproduced in a setting where we eliminated the control bounds on the cars (\figref{fig:s0unbounded}).

For model-free learning, we used the PPO2 implementation in the Stable Baselines repository \cite{baselines} with default parameters as our model-free algorithm. This method takes several orders of magnitude more than the model-based methods and is not able to match their performance. This can likely be attributed to the fact that the model-free method is not handed the dynamics of the system and therefore cannot take advantage of online planning, which is fundamental to the way model-free methods operate. Additionally, the accuracy of the controls required to successfully perform a merge make exploration challenging in this scenario, so the Gaussian noise used by PPO2 might also play a factor in the algorithm's relatively poor performance. 


\noindent\textbf{Takeaways.} Overall, what we find confirms intuition: if we have a good model, learning its parameters leads to good performance compared to learning from scratch. More surprising is the performance of the black-box methods, both off- and on-policy. Without seeing data of the human responding to the robot actually merging in front of them (which is what the idealized exploration brings), the black-box approach did not seem to be able to model this, and got stuck in the mode where the optimal solution is to go behind the human. In contrast, ToM could more easily make that leap from the data it was trained on. In our case, the pre-training already brought it most of the way there; in general though we expect that ToM can for instance take off-policy data of mild influence (when the robot is further away) and extrapolate it to stronger influence (when the robot is closer to the human) -- this seems to be difficult for the black-box model, judging by the off-policy black-box results. On the other hand, had we initialized ToM differently, and had the off-policy data not contain any influence, ToM would have looked the same as the on- and off-policy black box methods, and it is entirely possible that not even an on-policy ToM could have fixed that. Overall, the results speak mostly to the difficulty that black-box models might have in extrapolating from one pattern of interaction to another, because they lack that structure that ToM exploits.


\section{Performance under \\Incorrect Modeling Assumptions}
From what we have found in the previous section, Theory of Mind is appealing because it has low sample complexity and works from off-policy data. However, the bias introduced with this approach could lead to underfitting. To quantify this, we compare the ToM and model-based methods when we modify the human simulator. Because of the tremendous difference in terms of performance and sample complexity of the model-free method, we chose to omit it to focus on the aforementioned comparison.

\begin{figure}
    \centering
    \includegraphics[width=\columnwidth]{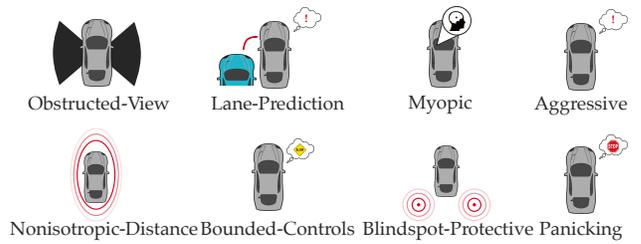}
    \caption{The various modifications made to the ground truth human simulator. The first row corresponds to modification of planning methodology, while the first three elements of the second row correspond to changes in reward features. The last corresponds to an irrational planning heuristic humans might use while under pressure.}
    \label{fig:mods}\vspace{-1em}
\end{figure}

\subsection{Human Simulators that Contradict Modeling Assumptions}

Inconsistency in how humans plan, the ``features" they might care about, or unexpected reactions to certain actions all violate the assumptions made by the ToM learner. We aimed to create ground truth modifications that are analogous to differences between reality and our ToM-based modeling -- things that designers of these systems might get wrong. As such, even though these are controlled experiments where we know the ground-truth, we think they provide some indication of how real-world differences would look like under different hypotheses. We group these modifications into 3 categories:
\begin{figure*}
    \centering
   \includegraphics[width=\textwidth]{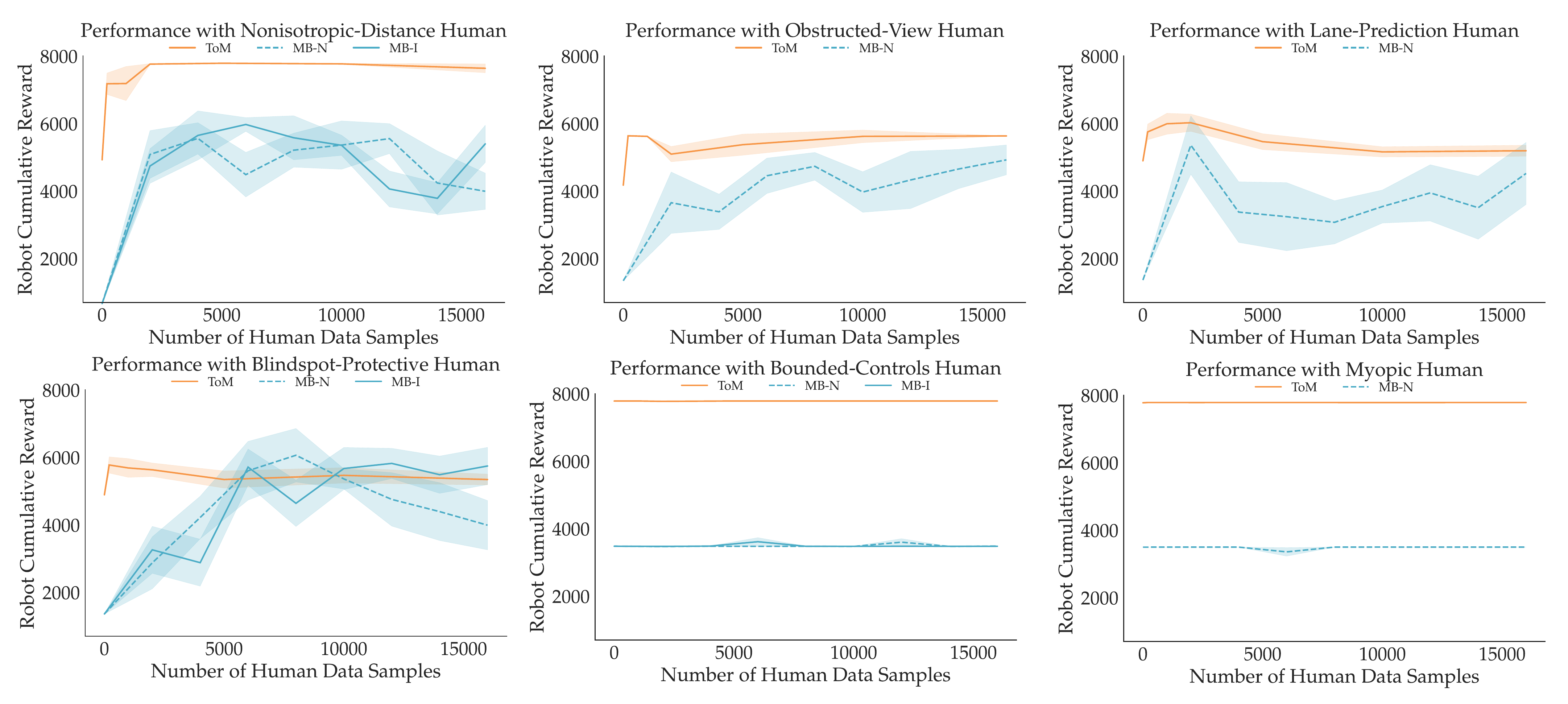}
    \caption{ ToM vs black-box model-based on different simulators. ToM is relatively robust to simulator modifications. This is in part due to the structure it is imposing being useful in low sample complexity regimes despite being wrong, but also because its predictions end up being wrong in a way that help compensate for the robot's planning suboptimality (see \figref{fig:agg}). } 
    \label{fig:simmods}
\end{figure*} 

\subsubsection{Incorrect model of how the human plans} One way our model could be wrong is if it inaccurately captures the planning process -- even if we assume the person is actually trying to optimize for a known reward, they might not optimize well or reason about the robot differently than ToM assumes.

\noindent\textbf{``Obstructed-View" -- Humans have blind spots.} Our instance of ToM assumes humans have a $360^\circ$ vision. In reality, drivers have blind spots. To model this we hide cars that are not in a double-cone from the human, with a vertex angle of 45$^{\circ}$.
Robots that do not model blind spots will thus take more risky maneuvers, expecting to be seen by the person when they are not.

\noindent\textbf{``Lane-Prediction" -- Humans can plan conservatively.} 
Given the inherent risks in driving, humans may be more cautious in their planning than necessary, taking evasive maneuvers when there is any chance of danger. This simulator swerves out of the current lane if the robot angles itself slightly towards said lane. A robot that does not know this might not be able to influence the human as much is possible. Note that this no longer matches our ToM's assumption that the human gets access to the robot's plan.

\noindent\textbf{``Myopic" -- Humans might not plan ahead for as long as the robot assumes.} 
Another assumption ToM makes about the human's reasoning is that it plans as far forward as the robot does. In reality, however, humans may be more myopic, and plan forward for a shorter time horizon than we assume. Our ``Myopic" human simulator only plans forward for one step.

\subsubsection{Incorrect model of what the human cares about} Another class of inconsistency in modeling deals with reward features. 

\noindent\textbf{``Nonisotropic-Distance" -- Humans care about avoiding other cars, but we might not know how sensitive they are to getting close to different areas of another car. } 
The original human simulator has a cost based on a Gaussian that takes in the Euclidean distance between the centers of the two cars, the ``Nonisotropic-Distance" simulator modifies the cost contours to be longer than they are wide, as show in Figure \ref{fig:mods}. This models the fact that some people are more comfortable with cars to their sides rather than in their own lane.


\noindent\textbf{``Bounded-Controls"-- We might not know people's preferences for speed or their control limitations. } 
Human drivers have different preferences for how fast they turn or accelerate as well as cars with different control bounds. To model this, the ground truth human simulator's values of $a_{max}$ and $\omega_{max}$ (see section 3.A) are reduced to $\frac{a_{max}}{2}, \frac{\omega_{max}}{2}$, reducing the capability of the human to react to the robot.

\noindent\textbf{``Blindspot-Protective" -- Humans might additionally care about not having another car in their blindspot.} 
A subclass of modifications we have not yet considered is where the human might use features in planning that the robot might not know about. One example of this might be discomfort with having cars in one's blindspot. Drivers might speed up or slow down to avoid such an arrangement. This modification models this dislike by adding additional points of Gaussian cost where blindspots are, as illustrated in \figref{fig:mods}.

 \noindent\textbf{``Panicking" -- Humans might be heuristic-driven.} 
 Humans might also use heuristics that are irrational. Our ``Panicking" modification combines the slower speed of the ``Bounded-Controls" modification with an additional heuristic of stopping immediately if another car is fewer than 2 car-lengths behind. This is inspired by a newer driver, whose inexperience causes him to drive slowly and panic when another car approaches. 

\subsubsection{A modification purposefully detrimental to the robot} Our last modification is meant to purposefully create dangerous situations when the robot is using the ToM assumption.
\noindent\textbf{``Aggressive" -- Humans might want to stay in front of autonomous cars.} 
New autonomous vehicles might seem dangerous to drivers and they may want to stay in front of them so they do not have to worry about false stops by the car. This modification is implemented by flipping the sign of the car-avoidance feature and moving the center of the Gaussian to a spot in front of the robot car.

\subsection{Analysis}

\begin{figure}
\includegraphics[width=\columnwidth]{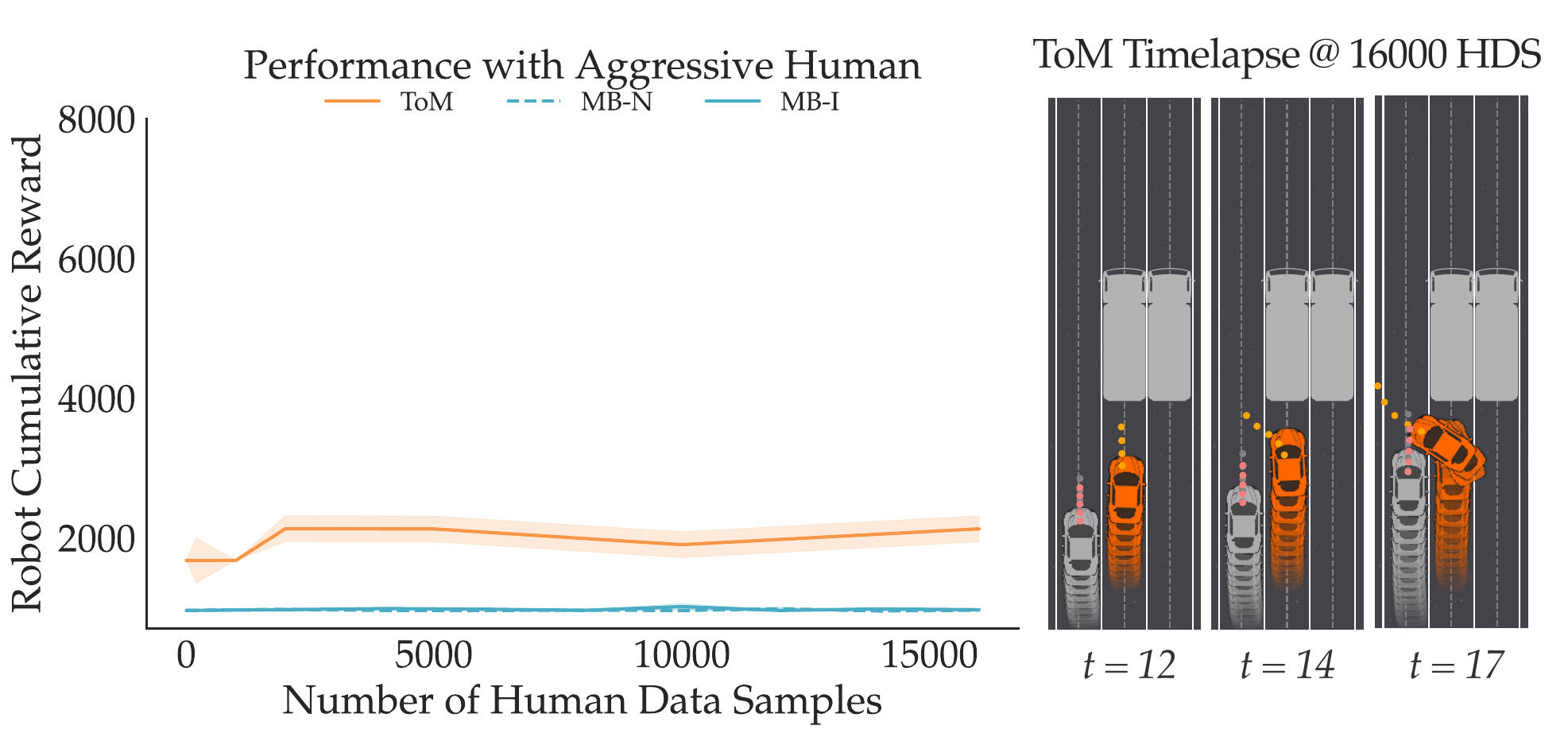}
     \caption{Rewards and sample trajectories for the aggressive human experiment. We see ToM doing better whereas black-box model based fails to improve the robot's reward as it gets more data. This is not so much a reflection on the predictions, but rather on the influence these predictions have on the planning output: ToM wrongly predicts the person would slow down so it deems optimal to merge in front when using a short horizon planner; this is not actually optimal for the short planing horizon when interacting with the "real" human, but leads to higher cumulative reward over the entire horizon.}
     \label{fig:agg}
 \end{figure}

 
\figref{fig:simmods} and \figref{fig:agg} plot the performance of ToM and black-box model-based\footnote{Panicking results were analogous to bounded-controls and myopic and were ommitted from the plot.}. For the latter, we used the idealized distribution when possible by getting the robot to plan with the ground truth modified model and collecting data. Additionally, we also collected data from the robot planning with the unmodified model ("MB-N" in the plots, which performs comparably). 

Overall, we see the ToM performing better initially across the board, and black-box catching up in most modifications with enough data. Surprisingly, we do not see black-box surpassing the performance of ToM. We found this to be caused by an interesting facet of planning in these environments, rather than by the predictions failing to improve. ToM also ends up with imperfect predictions, but the robot planning with it achieves higher reward nonetheless. This is because our planner uses a short horizon and is locally optimal. As the trajectories in \figref{fig:agg} show, the ToM method incorrectly predicts that the aggressive human will slow down more than they actually do. This fools the short horizon planner into deeming that merging in front of the human is the right option. While this is not actually right with respect to reward over that short horizon when interacting with the aggressive human, it ends up being the right decision for the long term, at least according to the reward function we specified. The robot ends up with an aggressive merge which accumulates more reward overall than the black-box model-based method. In a few of the modifications, the black-box approach failed to improve the robot's performance with more data, again because of the difference between prediction accuracy and planning with those predictions when using a suboptimal planner. There were also cases where ToM obtained maximum reward from the start, and that is because some modifications make the human slower and less aggressive, which enables the merge to happen seamlessly even from the start, when ToM operates solely based on the pre-training.

\noindent\textbf{Takeaways.} The ToM performance tends to quickly asymptote, and it typically does so at a higher reward that the black-box methods. This is in part because these misspecified ToM models make wrong predictions that help compensate, in our environment, for the planner's short horizon. While we expect that in general, when given data from the right distribution, black-box model-based asymptotes to higher robot reward than ToM, our finding speaks to the intricate relationship between prediction and planning, and how better predictions do not always lead to better plans. This is a subtle aspect that practitioners will need to consider in the design of algorithms for robots that interact with people.





\section{Transferability of Learned Models}

\begin{figure}
    \centering
    \includegraphics[width=\columnwidth]{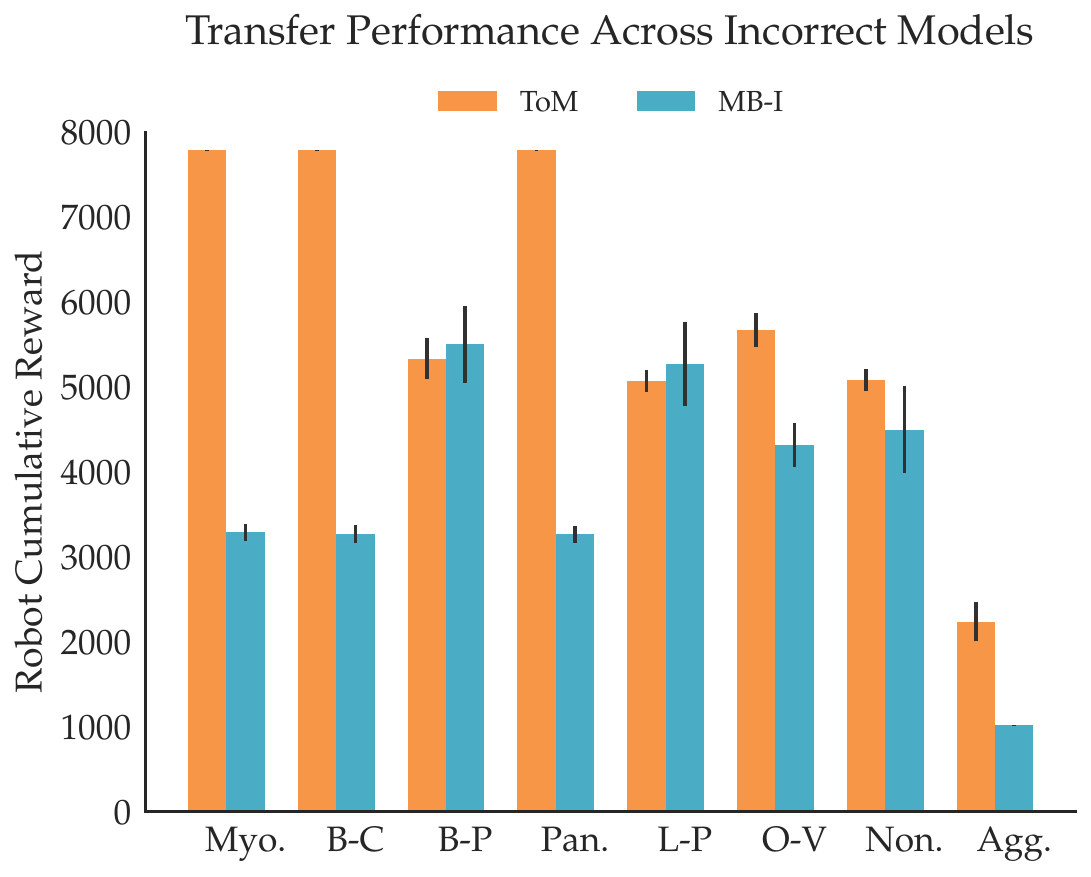}
    \caption{ Results from testing the models trained with the original human simulation in the modified simulations. Overall, ToM tends to transfer better.}
    \label{fig:transfer}
\end{figure}

Finally, we study the transferability of the models trained against the original human. We compare their performance when tested against the modified simulators from the previous section. \figref{fig:transfer} shows that all models transfer better when changes are small. 

Some modifications (myopic, bounded controls, and pankicking) make the human go slower in our environment, thereby making the merge that ToM attempts have high reward (even higher than with the original human). Despite the change in human behavior, ToM keeps making predictions that enable the robot to merge. On the other hand, the shift in the input distribution causes poor predictions from black-box model based, which in turn leads to poor performance. ToM also handles the aggressive and obstructed-view humans better. The blindspot-protective and lane-prediction humans behave similarly enough to the original human that both methods retain their original performance.

\section{Discussion}
\noindent\textbf{Summary.} We provided what is, to the best of our knowledge, the first comparison between model-free, black-box model-based, and Theory-of-Mind-based methods for interaction. We used a simple driving task to  quantify the performance advantage of ToM-based when we have made the right assumptions, as well its data collection advantage: it does not seem to require on-policy interaction data during learning, and can be trained on observed human-human data. 

We also found that black-box model-based methods can perform as well as ToM, but the training distribution matters: in our example, neither off- nor on-policy was successful, and instead the robot added exploration to on-policy via an idealized distribution (coming from the ground truth optimal policy). These methods did not seem able to extrapolate from mild effects of the robot on the human (in off-policy data) to strong effects (needed in the test environment). They also required much more data (an order of magnitude in our experiments). Further, we found that the model-free method required several orders of magnitude more data and was not able to achieve anywhere near the performance of the other methods considered.

We also studied what happens as ToM's assumptions become increasingly wrong by emulating the kinds of deviations we might expect to encounter. Lastly, we saw that ToM methods seem to transfer better.

\noindent\textbf{Limitations.}
Ultimately, our work does not answer the question of which type of model to use, it merely provides a few data points that are based on hypothetical human behavior. Our results are also for a particular domain, environment, and a particular ToM model. There are a few major caveats to these results as a consequence. First, the interaction with the planner suboptimality led to ToM actually benefit from its wrong assumptions. In other enviroments, the opposite might be true, i.e. that ToM might combine particularly poorly with a short planning horizon. On average, we expect that black-box model-based methods eventually surpass the ToM performance with enough data, unlike what happened in our environment. Second, even with perfect planners, more accurate prediction does not always mean better performance, which can again affect these results. In some environments, such as the robot starting behind the person and accelerating to move the person out of the way, random predictions are just as useful as accurate ones, and a black-box approach would perform really well from the start. 
Third, the environment also made it so that ToM could start with a very high performance just based on pre-training. It is not clear how much easier it is to pre-train ToM generally. Further, even though on-policy black-box model-based did not perform well in our environment, this can be addressed with better exploration without requiring the idealized distribution (at the cost of data and cumulative regret), or by mixing in relevant enough human-human interaction data. 

Overall, these results only scratch the surface of understanding the trade-offs with using Theory-of-Mind approaches for inductive bias in HRI. These trade-offs are not just based on sample complexity, but are instead complicated by the interaction between prediction accuracy and planning suboptimality, and by the feasibility of collecting on-policy data and of exploring. Looking forward, we are also excited to research \emph{hybrid} methods that might be able to take advantage of the best of each paradigm: using ToM assumptions in low-data regimes, having flexibility when there is enough data.








\section*{Acknowledgments}

We thank the members of the InterACT Lab at UC Berkeley. In particular, we are grateful for Kush Bhatia's feedback on building human simulators, Eli Bronstein's assistance on the black-box model-based component of this work, and Michael McDonald's work on parallelizing experiments.

This work is partially supported by NVIDIA, BDD, and the Caltech Arjun Bansal and Ria Langheim Summer Undergraduate Research Fellowship.

\bibliographystyle{ieeetr}
\bibliography{refs}

\begin{thebibliography}{10}

\bibitem{glascher2010states}
J.~Gl{\"a}scher, N.~Daw, P.~Dayan, and J.~P. O'Doherty, ``States versus
  rewards: dissociable neural prediction error signals underlying model-based
  and model-free reinforcement learning,'' {\em Neuron}, vol.~66, no.~4,
  pp.~585--595, 2010.

\bibitem{lee2014neural}
S.~W. Lee, S.~Shimojo, and J.~P. O’Doherty, ``Neural computations underlying
  arbitration between model-based and model-free learning,'' {\em Neuron},
  vol.~81, no.~3, pp.~687--699, 2014.

\bibitem{mnih2013playing}
V.~Mnih, K.~Kavukcuoglu, D.~Silver, A.~Graves, I.~Antonoglou, D.~Wierstra, and
  M.~Riedmiller, ``Playing atari with deep reinforcement learning,'' {\em arXiv
  preprint arXiv:1312.5602}, 2013.

\bibitem{schulman2015trust}
J.~Schulman, S.~Levine, P.~Abbeel, M.~Jordan, and P.~Moritz, ``Trust region
  policy optimization,'' in {\em International Conference on Machine Learning},
  pp.~1889--1897, 2015.

\bibitem{schulman2017proximal}
J.~Schulman, F.~Wolski, P.~Dhariwal, A.~Radford, and O.~Klimov, ``Proximal
  policy optimization algorithms,'' {\em arXiv preprint arXiv:1707.06347},
  2017.

\bibitem{finn2016unsupervised}
C.~Finn, I.~Goodfellow, and S.~Levine, ``Unsupervised learning for physical
  interaction through video prediction,'' in {\em Advances in neural
  information processing systems}, pp.~64--72, 2016.

\bibitem{WinNT}
C.~Perez, ``Predictive learning is the new buzzword in deep learning.''

\bibitem{gopnik199410}
A.~Gopnik and H.~M. Wellman, ``10 the theory theory,'' {\em Mapping the mind:
  Domain specificity in cognition and culture}, p.~257, 1994.

\bibitem{carruthers1996theories}
P.~Carruthers and P.~K. Smith, {\em Theories of theories of mind}.
\newblock Cambridge University Press, 1996.

\bibitem{gergely2003teleological}
G.~Gergely and G.~Csibra, ``Teleological reasoning in infancy: The na{\i}ve
  theory of rational action,'' {\em Trends in cognitive sciences}, vol.~7,
  no.~7, pp.~287--292, 2003.

\bibitem{sodian2004infants}
B.~Sodian, B.~Schoeppner, and U.~Metz, ``Do infants apply the principle of
  rational action to human agents?,'' {\em Infant Behavior and Development},
  vol.~27, no.~1, pp.~31--41, 2004.

\bibitem{baker2009action}
C.~L. Baker, R.~Saxe, and J.~B. Tenenbaum, ``Action understanding as inverse
  planning,'' {\em Cognition}, vol.~113, no.~3, pp.~329--349, 2009.

\bibitem{becker2013economic}
G.~S. Becker, {\em The economic approach to human behavior}.
\newblock University of Chicago press, 2013.

\bibitem{nikolaidis2013human}
S.~Nikolaidis and J.~Shah, ``Human-robot cross-training: computational
  formulation, modeling and evaluation of a human team training strategy,'' in
  {\em Proceedings of the 8th ACM/IEEE international conference on Human-robot
  interaction}, pp.~33--40, IEEE Press, 2013.

\bibitem{busch2017learning}
B.~Busch, J.~Grizou, M.~Lopes, and F.~Stulp, ``{L}earning {L}egible {M}otion
  from {H}uman--{R}obot {I}nteractions,'' {\em International Journal of Social
  Robotics}, pp.~1--15, 2017.

\bibitem{reddy2018shared}
S.~Reddy, S.~Levine, and A.~Dragan, ``Shared autonomy via deep reinforcement
  learning,'' {\em arXiv preprint arXiv:1802.01744}, 2018.

\bibitem{schmerling2017multimodal}
E.~Schmerling, K.~Leung, W.~Vollprecht, and M.~Pavone, ``Multimodal
  probabilistic model-based planning for human-robot interaction,'' {\em arXiv
  preprint arXiv:1710.09483}, 2017.

\bibitem{ziebart2009planning}
B.~D. Ziebart, N.~Ratliff, G.~Gallagher, C.~Mertz, K.~Peterson, J.~A. Bagnell,
  M.~Hebert, A.~K. Dey, and S.~Srinivasa, ``Planning-based prediction for
  pedestrians,'' in {\em Intelligent Robots and Systems, 2009. IROS 2009.
  IEEE/RSJ International Conference on}, pp.~3931--3936, IEEE, 2009.

\bibitem{dragan2013legibility}
A.~D. Dragan, K.~C. Lee, and S.~S. Srinivasa, ``Legibility and predictability
  of robot motion,'' in {\em Proceedings of the 8th ACM/IEEE international
  conference on Human-robot interaction}, pp.~301--308, IEEE Press, 2013.

\bibitem{bai2015intention}
H.~Bai, S.~Cai, N.~Ye, D.~Hsu, and W.~S. Lee, ``Intention-aware online pomdp
  planning for autonomous driving in a crowd,'' in {\em Robotics and Automation
  (ICRA), 2015 IEEE International Conference on}, pp.~454--460, IEEE, 2015.

\bibitem{javdani2015shared}
S.~Javdani, S.~S. Srinivasa, and J.~A. Bagnell, ``Shared autonomy via hindsight
  optimization,'' {\em arXiv preprint arXiv:1503.07619}, 2015.

\bibitem{sadigh-human-rss2016}
D.~Sadigh, S.~S. Sastry, S.~A. Seshia, and A.~D. Dragan, ``Planning for
  autonomous cars that leverage effects on human actions,'' in {\em Proceedings
  of Robotics: Science and Systems}, RSS '16, 2016.

\bibitem{tu2017least}
S.~Tu and B.~Recht, ``Least-squares temporal difference learning for the linear
  quadratic regulator,'' {\em arXiv preprint arXiv:1712.08642}, 2017.

\bibitem{abbeel04inverse}
P.~Abbeel and A.~Y. Ng, ``Apprenticeship learning via inverse reinforcement
  learning,'' in {\em ICML '04: Proceedings of the twenty-first international
  conference on Machine learning}, (New York, NY, USA), p.~1, ACM, 2004.

\bibitem{levineCIOC}
S.~Levine and V.~Koltun, ``Continuous inverse optimal control with locally
  optimal examples,'' {\em CoRR}, vol.~abs/1206.4617, 2012.

\bibitem{ziebart08maxent}
B.~D. Ziebart, A.~Maas, J.~A. Bagnell, and A.~K. Dey, ``Maximum entropy inverse
  reinforcement learning,'' in {\em Proceedings of the 23rd National Conference
  on Artificial Intelligence - Volume 3}, AAAI'08, pp.~1433--1438, AAAI Press,
  2008.

\bibitem{Andrew07L1Reg}
G.~Andrew and J.~Gao, ``Scalable training of l1-regularized log-linear
  models,'' in {\em Proceedings of the 24th international conference on Machine
  learning (ICML)}, (Corvalis, Oregon), pp.~33--40, 2007.

\bibitem{adam}
D.~P. Kingma and J.~Ba., ``Adam: A method for stochastic optimization,'' {\em
  arXiv preprint arXiv:1412.6980}, 2014.

\bibitem{ross2011reduction}
S.~Ross, G.~Gordon, and D.~Bagnell, ``A reduction of imitation learning and
  structured prediction to no-regret online learning,'' in {\em Proceedings of
  the fourteenth international conference on artificial intelligence and
  statistics}, pp.~627--635, 2011.

\bibitem{baselines}
A.~Hill, A.~Raffin, M.~Ernestus, A.~Gleave, A.~Kanervisto, R.~Traore,
  P.~Dhariwal, C.~Hesse, O.~Klimov, A.~Nichol, M.~Plappert, A.~Radford,
  J.~Schulman, S.~Sidor, and Y.~Wu, ``Stable baselines,'' 2018.

\end{thebibliography}

\end{document}